%% file: template.tex
\documentclass[smallextended,final,natbib]{svjour3}       
\smartqed  

\usepackage{graphicx}
 \usepackage{mathptmx}      
\usepackage[cmex10]{amsmath}
\usepackage{amsfonts}
\usepackage[nice]{nicefrac}
\usepackage{epsfig}
\usepackage{graphicx}
\usepackage{verbatim}
\usepackage{algorithm}
\usepackage{algorithmic}
\usepackage{bm}
\usepackage{subcaption}

\usepackage{float}

\usepackage{multirow}
\usepackage{array}
\usepackage[table]{xcolor}
\definecolor{lgray}{gray}{0.85}
\usepackage{booktabs}

\usepackage{arydshln}

\usepackage{tabularx}
\usepackage{natbib}
\usepackage{epstopdf}

\usepackage{tocbibind}
\usepackage[bookmarks]{hyperref}
\usepackage[title,titletoc,toc]{appendix}
\addcontentsline{toc}{section}{\refname}
\usepackage{xcolor}
\definecolor{dark-red}{rgb}{0.4,0.15,0.15}
\definecolor{dark-blue}{rgb}{0.15,0.15,0.4}
\definecolor{medium-blue}{rgb}{0,0,0.5}
\hypersetup{
  colorlinks   = true, 
  urlcolor     = blue, 
  linkcolor    = blue, 
  citecolor   = blue 
}

\begin{document}

\title{Network Traffic Decomposition for Anomaly Detection}



\author{Tahereh Babaie         \and
        Sanjay Chawla  \and \\
        Sebastien Ardon}


\institute{T. Babaie \at
              School of IT, University of Sydney,
Sydney, NSW, Australia \\
ATP Research Laboratory,
NICTA, Alexandria, NSW, Australia \\
              \email{tahereh.babaie@nicta.com.au}           
           \and
           S. Chawla \at
             School of IT, University of Sydney,
Sydney, NSW, Australia \\
              \email{sanjay.chawla@sydney.edu.au}
          \and
           S. Ardon \at
              ATP Research Laboratory,
NICTA, Alexandria, NSW, Australia\\
              \email{sebastien.ardon@nicta.com.au}
}
\date{Received: date / Accepted: date}

\maketitle

\begin{abstract}
In this paper we focus on the detection of network anomalies like
Denial of Service (DoS) attacks and port scans in a unified manner.
While there has been an extensive amount of research in network anomaly
detection, current state of the art methods are only able to detect one
class of anomalies at the cost of others. The key tool we will
use is based on the spectral decomposition of a trajectory/hankel matrix
which is able to detect deviations from both between and within
correlation present in the observed network traffic data.
Detailed experiments on synthetic and real network traces shows
a significant improvement in detection capability over
competing approaches. In the process we also address the issue of
robustness of anomaly detection systems in a principled fashion.
\keywords{Anomaly Detection \and Hankel Matrix \and SVD}
\end{abstract}

\input{introduction}
\input{dynamic3}

\input{mssa}

\input{anomalytype}
\input{evaluation}

\input{related}
\input{conclusion}

\begin{acknowledgements}
This work is partially supported by NICTA\footnote{http://nicta.com.au/}. NICTA is funded by the Australian Government as represented by the Department of Broadband, Communications and the Digital Economy and the Australian Research Council through the ICT Centre of Excellence program.
\end{acknowledgements}

\begin{appendices}
\section{Hankelization}
\input{hankelize}
\label{sec:hankelization}
\end{appendices}


\bibliographystyle{spbasic}

\bibliography{Mybib}

\end{document}

%% file: introduction.tex
\section{Introduction}
In its most abstract form, network traffic can be described
by a time series $y(t)$, where $y$ represents the observed state of
the traffic. For example, $y(t)$ could simply be the total number
of packets or could be a vector, where each component
represents an active flow. A flow is an aggregation of packets
by attributes like source and destination ip address.

In order to detect anomalies in network traffic we must first model the generative process, which
gives rise to the observable time series $<y(t)>$. Assume that the latent variables $x(t)$. The relationship between $y(t)$ and $x(t)$ can be abstractly represented by a model as $y(t)=f(x(t))$. We can learn the model and obtain an estimation as $\hat{y}=f(\hat{x}(t))$. Then an anomaly occurs of time $t$ if $y(t)-\hat{y}(t)$ is greater than a pre-defined threshold. In order to design the generative model we have to capture different forms of correlation between variables of the system which we describe here.


\subsection{Between and Within Flow Correlation}
\begin{figure} [t!]
\centering
\includegraphics[width=0.85\textwidth]{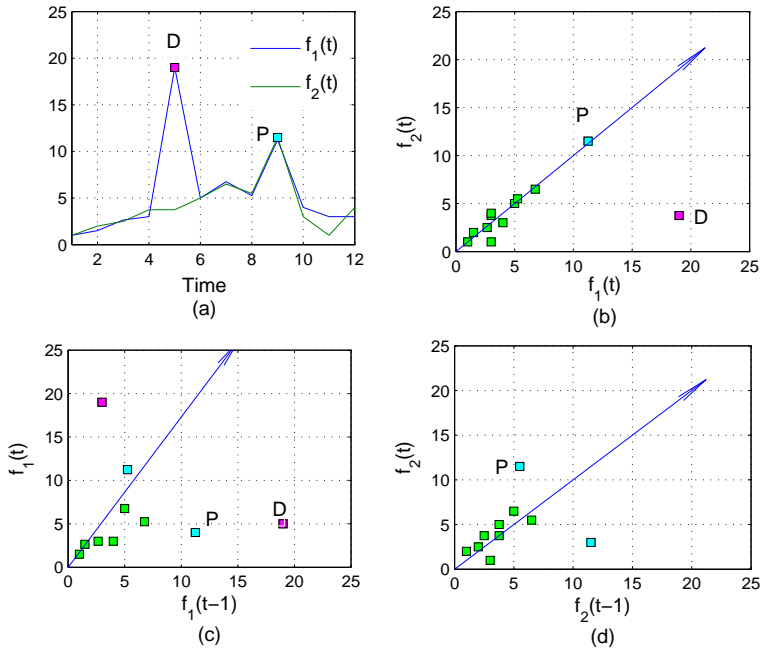}
\caption{(a) An example of two flows $f_1$ and $f_2$ experiencing two different anomalies DoS attack (D) and port scan (P). (b) SVD finds D anomaly and misses P one as it is in its normal space. (c) and (d): mapping the $f_1$ and $f_2$ vector into a 2-dimensional space and applying SVD both P and D anomalies are detectable.}
\label{fig:example3}
\vspace{-5pt}
\end{figure}

An important aspect that needs to be captured in any model of network
traffic is the presence of {\it between} and {\it within} correlation in
packet flows. For example, consider Figure \ref{fig:example3}(a), which
shows the the time series of two flows, $f_{1}(t)$ and $f_{2}(t)$.
The point labeled $D$ is an example  where the correlation {\it within}
flow $f_{1}(t)$ flows has deviated from the expected norm. Similarly, the point labeled $P$ is where the correlation {\it between} the two flows $f_{1}$ and
$f_{2}$  has deviated in a localized time window. The anomaly $D$ is
an example of a Denial of Service (DoS) attack while an anomaly $P$ is
an example of port scan. Discovering events like P and D is the focus of this paper.

\subsection{The Trajectory/Hankel Matrix}
A key tool that we will use to detect correlation deviation in network
traffic, is the trajectory (or Hankel) matrix that will be constructed
from the observed time series (see \cite{Takens1981,Broomhead1986a}). For example, given two
flows $\{f_{1}(i),f_{2}(i)\}_{i=1}^{T}$, the Hankel matrix ($H$) of window
length $L < T$ of the  two flows is given by \\[2ex]
\[
\left[\begin{matrix}
f_{1}(1) & \hdots & f_{1}(L) \\
f_{1}(2) & \hdots & f_{1}(L+1) \\
\vdots & \vdots & \vdots \\
f_{1}(T-L+1) & \hdots &f_{1}(T) \\
\end{matrix}
\right|
\left.\begin{matrix}
f_{2}(1) & \hdots & f_{2}(L) \\
f_{2}(2) & \hdots & f_{2}(L+1) \\
\vdots & \vdots & \vdots \\
f_{2}(T-L+1) & \hdots & f_{2}(T) \\
\end{matrix}
\right]
\]

Now the key insight of the paper, is that the SVD of correlation (or covariance) matrix of the Hankel matrix ($H$), will capture both {\it between} and
{\it within correlation} in network flows. Thus a low rank decomposition
of $H$ will characterize the manifold structure $M$ between the flows
as well as help identify the anomalies which deviate from the inferred
manifold structure. For example, Figure \ref{fig:example3}(b), shows
the relationship between the flows $f_{1}$ and $f_{2}$ and also the
direction of the most dominant eigenvector of the standard
correlation matrix (without the time lag). This decomposition is unable
to capture the port scan (P) anomaly because, $P$ is not a simple
violation of the between flow correlation but the existing correlation is
violated only in a localized time window. In Figure \ref{fig:example3}(c),
it is clear that a time window lag ($L=1$), captures the spatial
correlation in a small time window and thus the $P$ anomaly is
away from the main eigenvector. In Figure \ref{fig:example3}(d),
there is no correlation violation within flow $f_{2}$ and thus the
$P$ anomaly is in the direction of the main eigenvector.

The remainder of this paper is structured as follows. Section \ref{sec:ssa} explains the technique behind the singular spectrum analysis and its extension and compares both techniques with PCA. Section \ref{sec:evaluation} presents a validation of the different analysis algorithm based on SSA on a real traffic data and analyses their capability for anomaly detection. A brief background is presented in section~\ref{sec:related} and we discuss some conclusion remarks in section~\ref{sec:conclusion}.

%% file: dynamic3.tex
\section{Hankel Matrix and Generative Model}
\label{hankel}
We now justify the decomposition of the Hankel matrix based on a generative model of the data. In particular we
will show that if data is generated by a Linear Dyanmical System (LDS), then the SVD decomposition of the Hankel
matrix can be used to estimate the LDS.

\begin{figure}
\centering
\vspace{-15pt}
\includegraphics[width=0.7\textwidth]{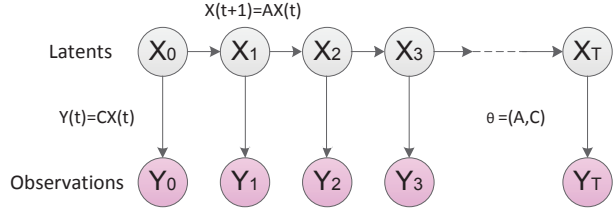}
\vspace{-20pt}
\caption{A linear dynamic latent model (LDS)}
\label{fig:LDS}
\end{figure}

Assume data is generated from a Linear Dynamical System (LDS) given by:
\[
\begin{matrix}
x(t + 1) & = Ax(t)+ w(t)  \\
y(t) & = Cx(t)+ v(t)
\end{matrix}
\]
where
\begin{itemize}
\item $x(t) \in \mathbb{R}^n$ is the system state vector,
\item $A$ defines the system's dynamics,
\item $w$ is the vector that captures the system error, e.g. a random vector from $\mathcal{N}(0,Q)$,
\item $y(t)\in \mathbb{R}^m$ is the observation vector,
\item $C)$ is the measurement function,
\item $v$ is the vector that represents the measurement error, e.g. a random vector from $\mathcal{N}(0,R)$,
\end{itemize}
Fig.\ref{fig:LDS} presents a graphical model of LDS.

\begin{problem} Assume that data is generated from an LDS governed by the equation above. Given
a sequence of observations $\{y_{i}\}_{i=1}^{n}$, estimate $A,C,Q$ and $R$.
\end{problem}

To solve the above problem, we need to define the Hankel matrix of the observations as
\[
H(t)= \begin{pmatrix}
y(t) & y(t+1) & y(t+2) & ... & y(n-\ell+1)\\
y(t+1)& y(t+2) & \ddots &  & \vdots \\
\vdots &&&&\\
y(t+\ell) & \ldots & & & y(n)\\
\end{pmatrix}
\]

\noindent
where y(t) is $m \times n$ observation at time $t$, and $H$ is a $\ell \times n'$ where $n'=n-\ell +1$.
Equivalently, $H$ is a Hankel matrix if and only if there exists a sequence ,$s_1,s_2,...$ such that $H_{i,j}=s_{i+j-1}$ (see \cite{Iokhvidov1982}). Therefore, every Hankel matrix uniquely determines a time series and every time series can be transferred into a Hankel matrix, i.e.:

\[
H(t-i) \Leftrightarrow y^i(t)
\]

\noindent
where $y^i(t)=\{y(i), y(i+1),...,y(t),...\}$. By replacing the entries of the Hankel matrix with their equivalent from the LDS:

\[
\small{
H(1)= \begin{pmatrix}
CAx(0) & CAx(1) & CAx(2) & ... & CAx(n-\ell)\\
CAx(1) & CAx(2)& \ddots &  & \vdots \\
\vdots &&&&\\
CAx(\ell-1)& \ldots & & & CAx(n-1)\\
\end{pmatrix}
 =\begin{pmatrix}
CAx(0) & CA^2x(0) & CA^3x(0) & ... & CA^{n-\ell+1}x(0)\\
CA^2x(0) & CA^3x(0)& \ddots &  & \vdots \\
\vdots &&&&\\
CA^{\ell}x(0)& \ldots & & & CA^nx(0)\\
\end{pmatrix}
}
\]
\[
=
\begin{array}{rll}
\begin{pmatrix}
CA && CA^2 && CA^3 & ... && CA^{\ell}\end{pmatrix}^T & \cdot &
\begin{pmatrix}
x(0)& &Ax(0)&& A^2x(0)&&\ldots && A^{n-\ell-2}x(0)\\
\end{pmatrix}
\end{array}
\]

\noindent
Define:
\[
\begin{array} {l}
P=\begin{pmatrix} CA & & CA^2 & &CA^3 & ... & &CA^{\ell}\end{pmatrix}^T \\
Q=\begin{pmatrix} x(0)& & Ax(0)& &A^2x(0)&\ldots & &A^{n-\ell-2}x(0) \end{pmatrix}
\end{array}
\]

\noindent
then:
\[
\begin{array}{rl}
H(1)=& PQ
\end{array}
\]

\noindent
The shifted Hankel matrices can be described by:
\[
\begin{array}{rl}
H(i)=& PA^{i-1}Q
\end{array}
\]

\noindent
To obtain the matrices $A$ and $B$, perform singular value decomposition of $H(1)$:

\[
\begin{array} {rl}
H(1)& =  U \Sigma^2 V^T \\
\end{array}
\]

\noindent
where $\Sigma^2$ is a diagonal $\ell \times \ell$  matrix containing the singular values and the $\ell$ columns of $U$ are the singular vectors. Selecting the top-k $(1< k < \ell)$ singular values from the matrix $\Sigma^2$,denoted by $\Sigma_k$, and $k$ associated singular vectors, denoted by $U_k$, we define reduced rank matrices:

\[
\begin{array}{rl}
P_k \doteq & U_k \Sigma_k
\end{array}
\]
\[
\begin{array}{rl}
Q_k \doteq &  \Sigma_k V^T
\end{array}
\]

\noindent
Using the 1-shifted Hankel matrix $H(2)$ and the reduced rank matrices $P_k$ and $Q_k$:

\[
\begin{array} {rl}
H(2)& = P_kA_kQ_k \\
    & = U_k \Sigma A_k\Sigma V^T \\
\end{array}
\]

\noindent
Then the matrix $A_k$ can be approximated as:
\[
\begin{array} {rl}
A_k & = (U_k \Sigma_k)^{-1} H(2) (\Sigma_k V^T)^{-1} \\
\end{array}
\]

\noindent
Then, given $A_k$ we can estimate $C_k$ as:
\[
\begin{array} {rl}
C_k & = P_1^{-1} A_k \\
\end{array}
\]

\noindent
where $P_1$ is the first $m$ rows of the matrix $P$.
Given $A_k$ and $C_k$, we can estimate
\[
\begin{array} {rl}
\delta_k=& y-\hat{y}\\
        =& y-C_k\hat{x}
\end{array}
\]

\noindent
An outlier is reported whenever $|\delta_k|$ exceeds a predefined threshold.\\

In practice, we are able to use the decomposition of the Hankel matrix to identify outliers. Recall once again the SDV of the Hankel matrix:
\[
\begin{array} {rl}
H(1)& =  U \Sigma^2 V^T \\
    & = \sum _{i=1}^{k} \lambda_i^{1/2}U_iV_i' + \sum _{i=k+1}^{\ell} \lambda_i^{1/2}U_iV_i' \\
\end{array}
\]

\noindent
If we define $\hat{H} \doteq \sum _{i=1}^{k} \lambda_i^{1/2}U_iV_i'$ and $\Delta_k \doteq \sum _{i=k+1}^{\ell} \lambda_i^{1/2}U_iV_i'$ then:
\[
\begin{array} {rl}
\Delta_k & = H(1) - \hat{H} \\
\end{array}
\]

\noindent
We know that every Hankel matrix is associated with a time series. Therefore if these matrices would be Hankel then we can obtain the error space. This can be performed by means of diagonal averaging procedure. The averaging over the diagonals $\mathrm{i+j=const}$ of a matrix is called Hankelization. It transforms an arbitrary $\ell \times n'$ matrix to the form of a Hankel matrix, which can be subsequently converted to a time series. A Detailed procedure of Hankeliztion is given in Appendix \ref{sec:hankelization}.

%% file: mssa.tex
\section{Multivariate Singular Spectrum Analysis}
\label{sec:ssa}
The application of SVD to Hankel matrix is known as SSA or M-SSA.
The key advantage of M-SSA is its ability to succinctly capture both between (spatial) and within (temporal) correlation
in the underlying network traffic flows. Here we give a step-by-step introduction to
SSA, as a method of discovering anomalies.
\begin{enumerate}
\item
Assume the network flow volume through a router at a pre-specified level of granularity (e.g.five minutes) is given by the time series.
\[
y_{1},y_{2},\ldots,y_{m},w_{m+1},w_{m+2},\ldots,w_{n},y_{n+1},y_{n+2},\ldots
\]
We have used both $y$ and $w$ to indicate that the nature
of traffic has changed for $n-m+1$ time steps after $y_{m}$. In practice we
of course don't know where and when the traffic changes and is precisely what
we want to infer.\\
\item
Choose an integer $\ell < m$, known as the embedding dimension and
form the {\em Hankel matrix} for the $x$ part of the time series.
\[{\bf Y} =
\left(
\begin{array}{llll}
y_{1} & y_{2} & \ldots & y_{\ell} \\
y_{2} & y_{3} & \ldots & y_{\ell+1} \\
\ldots  & \ldots & \ldots & \ldots  \\
y_{m-\ell+1} & y_{m-\ell +2} & \ldots & y_{m}
\end{array}
\right)
\]
Where each ${\bf Y_{i}} = (y_{i},y_{i+1},\ldots,y_{i+\ell})'$, is of dimension $\ell$.  In SSA, the assumption is that ${\bf Y}$ captures the main
dynamics of the network flow. We now apply the Singular Value Decomposition (SVD) of ${\bf Y}$ as follows.\\
\item
For the $\ell \times \ell$ covariance matrix of $Y$ give by
\[
C = Y \times Y'
\]
\vspace{1pt}
\item
Compute the eigendecomposition of $C =[U,D]$ where $U$ is matrix
where each column is a eigenvector and $D$ is the diagonal matrix
of eigenvalues. The relationship between $C$, $U$ and $D$ is given
as
\[
CU(:,i) = D(i,i)U(:,i) \mbox{ for each $i$ }
\]
\vspace{1pt}
\item Form an $k$-dimensional  subspace $M$ of $R^{\ell}$ where $k \leq \ell$, by using the top-k eigenvectors of $U$, i.e., ${\bf M} = U_{s}U_{s}'$.
The space ${\bf M}$ is where the ``normal'' traffic lives and our objective
is to look for changes in the flow which cannot be explained by
${\bf M}$. This is achieved by projecting a sliding window of
$\ell$ dimensional vectors on $M$ and raising an alarm whenever
the deviation between a vector and its projection on $M$ becomes large.
\vspace{4pt}
\item
For example, consider a $\ell$-dim vector which contains parts of the
changed traffic $y_{i}'s$.
\[
{\bf z} = (y_{m-1},y_{m},w_{1},\ldots, w_{\ell - m -1})'.
\]
Then, the deviation between ${\bf z}$ and its projection on $M$
is given by ${\bf e} = \|{\bf z -  Mz}\|$.  Assuming that the $w_{i}$'s were generated by anomalous traffic, then the deviation ${\bf e}$ will be large relative to deviations caused by normal traffic.\\
\item To reconstruct the refined time series we proceed in a manner inverse to the step 2. On the
other hand, if the objective is to reconstruct the original time series
then we have to apply a hankelization (inverse) operator. The network anomaly
detection process remains unaffected by the inverse operation. More details
can be found in ~\cite{Vautard1989,Ghil2002,Golyandina2010}.
\end{enumerate}

\begin{figure} [t!]
\centering
\includegraphics[width=0.7\textwidth]{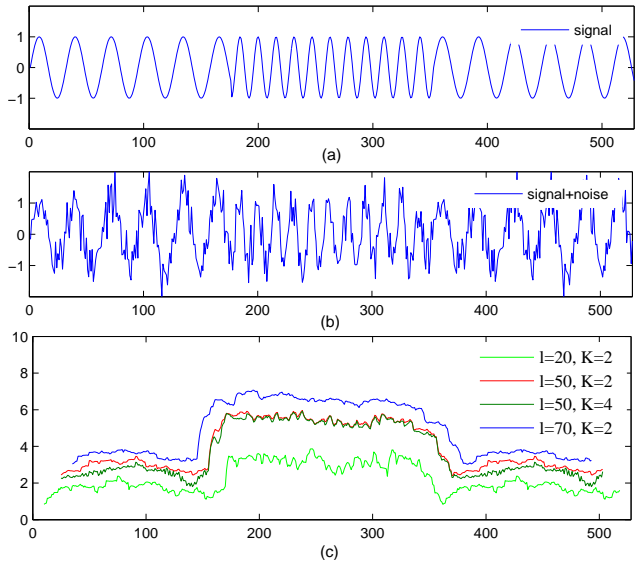}
\caption{An example of using SSA to detect changes in a time series for
various combination of parameter values $\ell$ and $k$. The time
series changes in the middle which is reflected in the deviation
in the bottom figure.}
\label{fig:example}
\end{figure}

\noindent
Before we go into further details about SSA we illustrate the key steps using a simple example.
\begin{example}Assume that a sample time series
is given as
\vspace{0.8pt}
\[
{\bf y(t)} =
\left\{
\begin{array}{ll}
sin(.2t) + \varepsilon(t) & \textit{if} \quad  1 \leq t \leq 175 \\
sin(.3t) + \varepsilon(t) & \textit{if} \quad 176 \leq t \leq 375 \\
sin(.2t) + \varepsilon(t) & \textit{if} \quad 376 \leq t \leq 560 \\
\end{array}
\right.
\]
\vspace{0.8pt}
Here $\varepsilon(t)$ is gaussian $\mathcal{N}(0,1)$ noise. Notice that there is
a change in the time series between $t=176$ and $t=376$.
Fig.~\ref{fig:example}(a and b) show the example time series without
the noise and the time series with added noise. Fig.~\ref{fig:example}(c)
shows the deviation of the signal for different values of $\ell$ and $k$.
It is clear that the deviation becomes larger near time step $176$
and then returns to its normal value after the change signal
disappears around time step $376$.
\end{example}
\subsection{Choice of Parameters in SSA}
The key idea in SSA is the use of a trajectory matrix ${\bf Y}$
which then factorized using SVD. The formal relationship between
$Y$ and the underlying dynamics of the time series has been
extensively researched in both the statistics and physics community.
The key take away from the theoretical literature is that for
an appropriate choice of $\ell$, the trajectory matrix will capture
the appropriate dynamics of the underlying system (see \cite{Takens1981,Broomhead1986a,Broomhead1986b,Broomhead1986c}). The choice of
$\ell$ along with $k$ (the dimensionality of the projected subspace)
and the  threshold $({\bf e})$ are three important parameters
that need to calibrated and set. These parameters are like ``knobs''
which a network administrator can use to adapt to specific
network characteristics.

Example 1 above already provides some indication of how the choices
of $\ell$ and $k$ have on time series monitoring. For example, for $\ell=20$,
the deviation ${\bf e}$ is less than for other values of $\ell$. This
may surprising at first but notice the initial part of the time series
has an intrinsic dimensionality of $1$ (as it is composed of one $\sin$ term).
Thus a smaller value of $\ell$ is better at capturing the dynamics
of the time series than a larger value $\ell=50,70$. Now consider,
the two cases where $L=50$ but $k=2$ or $k=4$. Notice that the projected
error (in the middle) is almost identical but at the tails the projection
error is higher for $k=2$ than $k=4$. This shows that while the choice
of $k$ has a significant impact on the projection error of the normal
traffic, when it comes to detecting the anomalous part the method
is quite robust for different choices of $k$. In fact this is one
of the key strengths of SSA that we will exploit in the analysis of
real network traffic data.

%% file: anomalytype.tex
\section{Network Anomaly Types}
\label{sec:atypes}
A key contribution of our paper is that the approach based on M-SSA
is able to detect almost all known types of network anomalies. In
this section we describe the different types of common
anomalies and explain why M-SSA provides  subsumes
other anomaly detectors. 
Table \ref{table:type} lists the common anomalies defined using the flow as a 5-tuple (source IP address, destination IP address, source port number, destination port number, transport protocol). More details can be found in \cite{Silveira2010a, Silveira2010b, Lakhina2005,Lakhina2004b}.\\

\noindent
A \textbf{Denial of Service (DoS)} attack occurs when the  attacking hosts send a large number of small packets - typically TCP SYN segments - to the attacked host and service, i.e. a single IP address and port number, in
order to deplete the system resources in the target host. The resulting traffic from DoS attack consists of a relatively small number of flows with  large packet counts as DoS attack tools often forge the source port number. Note that the specific case of Distributed Denial of Service (DDoS) attacks is effectively the same attack, but with several source IP addresses. The number of attacking hosts however, is typically much smaller than the packet count. We thus consider DDoS to be a special case of a DoS attack, and label as such.\\

\noindent
\textbf{Port scans} are typically used by attackers to discover open ports on the target host. This is accomplished by sending small packets as
connections requests to a large number of different ports on a single
destination IP address. At the flow level, they are therefore characterized
as an increase in the number of flows, each with a small packet count.\\

\noindent
\textbf{Large file transfers} are characterized by a few flows with packet counts
which are significantly larger than what common applications use. \\

\noindent
\textbf{Prefix outages} occurs when part of the network becomes unreachable, they can be identified when traffic from one or more IP prefixes disappears, which translates in a drop in the number of flows.\\

\noindent
\textbf{Link outage} is in a way a more severe version of Prefix outage, where the number of flows on the link drop close to zero.

\begin{table} [t!]
\caption{Network anomalies considered}
\centering
\begin{tabularx} {\linewidth}{l p{250pt}}
\toprule
Anomalies  & Description (flow is defined as one 5-tuple)  \\ [0.5ex]
\midrule
\textbf{DoS attack} &  a few flows with a large increase in packet count \\
\cellcolor{lgray}\textbf{port scan}   & \cellcolor{lgray} large increase in number of flows with a small packet count\\ [0.3ex]
\textbf{large file transfer}  & a few flows with a large increase in packet count, (but typically less than DoS attack)\\ [0.3ex]
\cellcolor{lgray}\textbf{prefix outages}  & \cellcolor{lgray} drop in number of flows (from one IP prefix) \\ [0.3ex]
\textbf{link outages}  & time intervals where all traffic disappear.\\ [1ex]
\bottomrule  
\end{tabularx}
\label{table:type}
\end{table}

%% file: evaluation.tex
\section{Experimental evaluation}
\label{sec:evaluation}
We have evaluated our proposed approach using both real and synthetic data
sets. For comparison we have implemented well known network anomaly
techniques based on wavelets, kalman filtering, fourier analysis
and the more recent ASTUTE method. The use of
synthetic data sets and simulation is a prerequisite for a rigorous evaluation strategy for network anomaly detection (\cite{Ringberg2008,Soule2005,Silveira2010a}).

\begin{table} [ht]
\caption{Alternative methods used in the experiments}
\centering
\begin{tabularx} {\linewidth} {p{300pt} r}
\toprule
Techniques are implemented by adjusting parameters as proposed in the literature. \\ [0.5ex]
\midrule
\textbf{Fourier analysis} \cite{Zhang2005} &  \\
\multicolumn{2}{p{320pt}}{We use fast Fourier transform (FFT) algorithm and set the cut off frequency to one cycle per 2 hours.} \\ [4ex]
\textbf{Wavelet analysis} \cite{Barford2002,Zhang2005} &   \\
\multicolumn{2}{p{320pt}}{We use a multi-level, 1-dimension wavelet algorithm, with Daubechies mother wavelet of order 6 and set the cut off frequency to 3.}  \\ [4ex]
\textbf{Kalman Filter} \cite{Soule2005}& \\
\multicolumn{2}{p{300pt}}{The target false positive rate of $2 \times 10^{-5}$ is applied to the innovation process.} \\ [1.5ex]
\textbf{ASTUTE} \cite{Silveira2010a,Silveira2010b}&  \\
\multicolumn{2}{p{300pt}}{The target false positive rate of $2 \times 10^{-5}$ is applied to the $AAV$ process.} \\ [1ex]
\bottomrule 
\end{tabularx}
\label{table:methods}
\end{table}

\subsection{Detection Capability}
We evaluate the detection capability of M-SSA using two real network
traces which we now describe.

\subsubsection{Datasets}
The first traffic trace if from the  Abilene network\footnote{Internet2 - http://www.internet2.edu/} and has
been used previously for network anomaly detection (see \cite{Silveira2010a, Silveira2010b, Lakhina2005,Lakhina2004b}). The data set consists of a one month traffic trace from a backbone router in New York during August 2007.
The Juniper router used to collect the data generated sampled J-flow statistics at the rate of 1/100. The flows were aggregated at
five minute intervals. The key attributes of the flow are: number of packets, number of distinct source
IP addresses, number of distinct destination IP addresses, number of
distinct source port numbers and number of distinct destination ports numbers.

The second, and more recent, traffic trace is from the MAWI (Measurement and Analysis on the WIDE Internet) archive project in Japan\footnote {http://www.wide.ad.jp/project/wg/mawi.html}. Here the data was sampled from a 150Mbps trans-pacific link between Japan and the United States for 63-hours in April 2012.

Labelling traffic traces with anomalies is notoriously difficult. The commonly accepted method is to combine algorithmic detection with manual inspection of the data. We have followed the URCA (\emph{Unsupervised Root Cause Analysis})
method proposed by ~\cite{Silveira2010c} with a false positive rate of $2 \times 10^{-9}$, followed by a thorough manual inspection of the data set.

\subsubsection{Results}
Table \ref{table:result} and Fig.~\ref{fig:result} show the results of the different
methods including M-SSA. The following are the key take aways.

\begin{enumerate}
\item
M-SSA is capable of detecting
a much wider range of anomalies regardless of their types. For the
Abilene data, M-SSA was able to identify 100\% of DoS attacks and
over 95\% port scans. Similarly on the MAWI data set the detection
rate was 100\% for DoS attacks and over 90\% for port scans.
\item
All other techniques (which were compared) can be placed in two groups:
Wavelets, Kalman and Fourier have high detection rates only for DoS attacks
while ASTUTE performs exceedingly well only for port scan anomalies.
\item In the Abilene data, around 7\% of the anomalies are related
to link outages. Here again, M-SSA has a 100\% detection rate and
except for Fourier, other techniques also have a high detection rate
with Wavelets doing the best.
\end{enumerate}

\begin{table}
\caption{Number of anomalies per type found by each technique in two traffic traces from Abilene and WIDE networks. M-SSA is able to discover both DoS and port scan in both networks.}
\centering
\resizebox{11cm}{!}{
\begin{tabularx} {376pt}{ >{\bfseries \sffamily}r  >{\sffamily}r  >{\sffamily}c  >{\sffamily}l  >{\sffamily}c >{\sffamily}c  >{\sffamily}c >{\sffamily}c >{ \bfseries \sffamily}c >{ \sffamily}c  >{\sffamily}c >{\sffamily}c}
\toprule
\addlinespace[5pt]
\multicolumn{11}{l}{\textsf{\textbf{\fontsize{9pt}{1em}\selectfont Trace: Internet2, from Abilene backbone}}}\\ [0.8ex]
\multicolumn{11}{l}{\textsf{Period: August 2007}}\\ [0.8ex]
  & &   & & ASTUTE &  Kalman & Wavelet &  Fourier & M-SSA & Hybrid$^\dag$ \\ [0.99ex]
\midrule
Anomalies class   & & Labeled  & &  &   &  &   & & \\
\addlinespace[7pt]
 \cellcolor{lgray} DoS attacks & & \cellcolor{lgray} 44 && \cellcolor{lgray}1 & \cellcolor{lgray}37 & \cellcolor{lgray} 41& \cellcolor{lgray}17& \cellcolor{lgray}44 & \cellcolor{lgray} 44 \\ [0.8ex]

 port scans  & & 221 && 198 & 0  & 18 & 0& 211& 216\\ [0.8ex]

 \cellcolor{lgray}large-file transfer & &\cellcolor{lgray} 2 && \cellcolor{lgray}2  & \cellcolor{lgray}0  & \cellcolor{lgray}0 & \cellcolor{lgray}0 & \cellcolor{lgray}2& \cellcolor{lgray} 2 \\ [0.8ex]

 link outage  & & 18 && 12 & 12  & 17 & 6&18& 18\\ [0.8ex]

 \cellcolor{lgray} prefix outage  & & \cellcolor{lgray} 1 &&\cellcolor{lgray} 1 &\cellcolor{lgray} 0  & \cellcolor{lgray}0 &\cellcolor{lgray} 0& \cellcolor{lgray}1 & \cellcolor{lgray} 1\\ [0.8ex]

Total found & &276 && 214  & 51  & 76 & &  265 & 271 \\ [6ex]
\toprule
\addlinespace[6pt]
\multicolumn{11}{l}{\textsf{\textbf{\fontsize{9pt}{1em}\selectfont Trace: MAWI, from WIDE backbone}}}\\ [0.8ex]
\multicolumn{11}{l}{\textsf{Period: April 2012$^\ddag$}}\\ [0.8ex]
  & &   & & ASTUTE &  Kalman & Wavelet &  Fourier & M-SSA & Hybrid$^\dag$ \\ [0.5ex]
\midrule
Anomalies class   & & Labeled  & &  &   &  &   & & \\
\addlinespace[7pt]
 DoS attacks & & 9 && 1 & 7 &  8 & 4& 9 & 9 \\ [0.8ex]

 \cellcolor{lgray}port scans  & & \cellcolor{lgray}98 && \cellcolor{lgray}89 & \cellcolor{lgray}11  & \cellcolor{lgray}19 & \cellcolor{lgray}0 & \cellcolor{lgray}89 & \cellcolor{lgray} 89\\ [0.8ex]

 large-file transfer  & &  1 && 1  & 1  & 0 & 0 & 1&  1 \\ [0.8ex]

 \cellcolor{lgray}link outage  & & \cellcolor{lgray}2 && \cellcolor{lgray}2 & \cellcolor{lgray}1  & \cellcolor{lgray}1 & \cellcolor{lgray}0 & \cellcolor{lgray}2 & \cellcolor{lgray}2\\ [0.8ex]

Total found & &111 && 93  & 20  & 28 & 4 &  101 & 101 \\ [4ex]
\bottomrule
\multicolumn{10}{l}{$^\dag$ Hybrid refers to ASTUTE $\cup$ Kalman $\cup$ Wavelet } &  \\
\multicolumn{10}{l}{$^\ddag$ This a 63-hours trace in the early days of the month. } &  \\
\end{tabularx}
}
\label{table:result}
\end{table}

\begin{figure}
        \centering
        \begin{subfigure}[b]{0.8\textwidth}
                \centering
                \includegraphics[width=\textwidth]{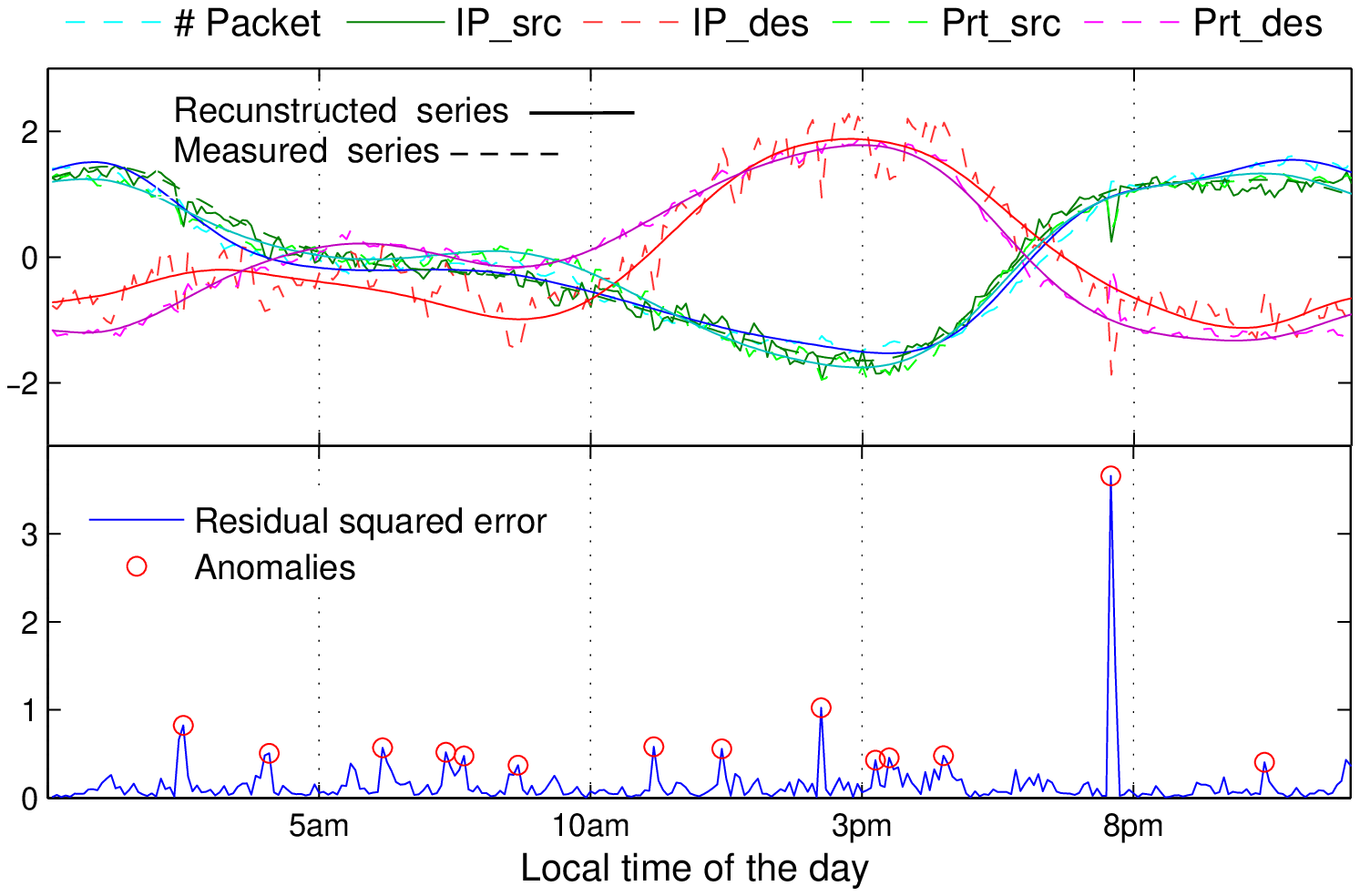}
                \caption{Internet2 traffic}
                \label{fig:gull}
        \end{subfigure}%
        ~ 

        \begin{subfigure}[b]{0.8\textwidth}
                \centering
                \includegraphics[width=\textwidth]{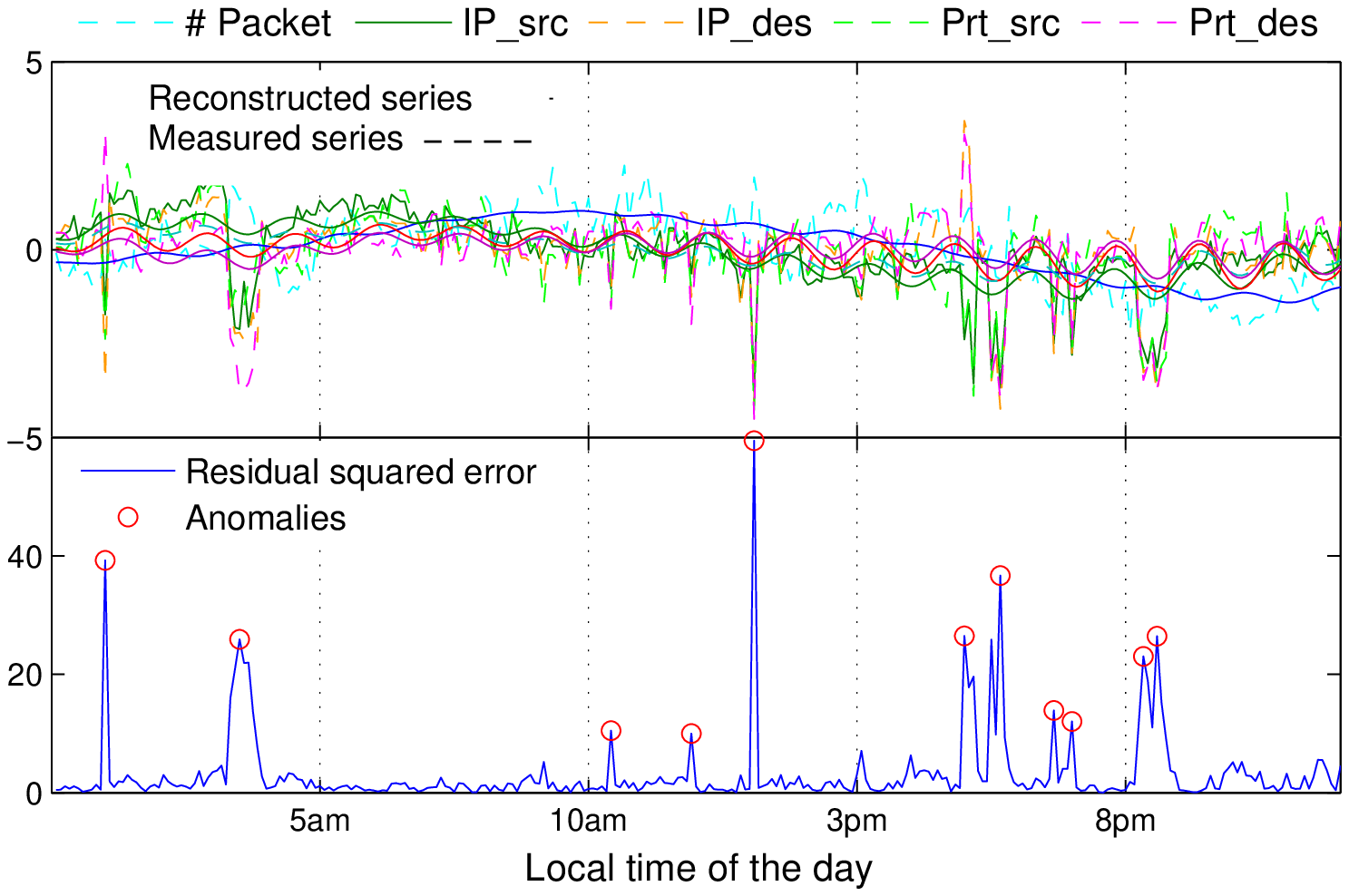}
                \caption{MAWI traffic}
                \label{fig:tiger}
        \end{subfigure}
        ~ 

        \caption{Timeseries plots of measured and reconstructed data along with related residual vector squared magnitude; for one day of both traffic traces from Abilene and WIDE networks. Triggered alarms shown as red circles.}
        \label{fig:result}
\end{figure}
\begin{figure} [t!]
\centering
\includegraphics[width=0.8\textwidth]{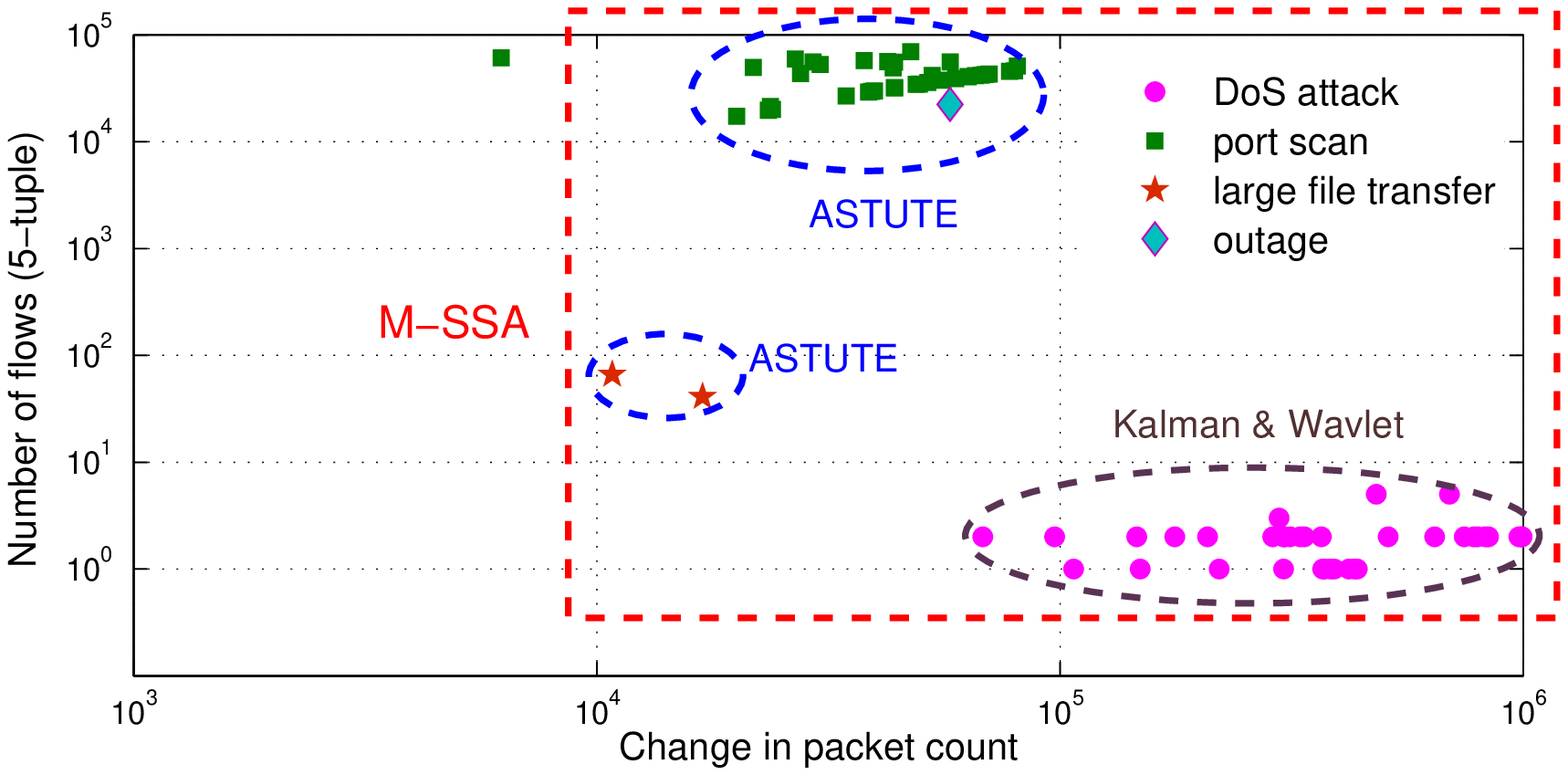}
\caption{Anomalies feature map shows DoS attacks are associated with a small number of flows with large number of packets, while port scans are a larger number of flows correlated in same time. The coverage of M-SSA subsumes all the techniques.}
\label{fig:Char}
\vspace{-10pt}
\end{figure}

To understand the results better we have carried out a deeper analysis
by examining the characteristic features of the anomalies.
In Fig.~\ref{fig:Char} we plot the known
Abilene anomalies using two features. The x-axis represents the change in packet
counts between two consecutive time bins. The y-axis represents the number of
distinct flows (5 tuples) in the time bin.

The first observation is that the set of anomalies are clustered in distinct groups, with the set of anomalies detected by Wavelet and Kalman approximately common (Wavelet is slightly better in detecting some port scans). Secondly, the Kalman filter and Wavelet techniques are not able to find anomalies caused by large number of flows with small packet counts. These includes anomalies where the rate of change in packet count in individual flows over time is small, e.g.
port scans, prefix outages and file transfers.  Wavelet as a time-frequency technique is able to flag sudden changes in traffic, but will miss any small variations such as port scans and absorb them in the main trend.

The Kalman filter technique is effective at detecting anomalies when the packet count variation over time is significant, such as DoS attacks. This is expected, as Kalman Filtering is essentially a forecasting technique in the time dimension. Another observation is that {\footnotesize ASTUTE} is not able to detect anomalies involving a few large flows (bottom right hand corner of Fig.~\ref{fig:Char}), such as DoS attacks. This is also expected, as {\footnotesize ASTUTE} is not able to detect large volume change in a few number of flows, because the $AAV$ process threshold is not violated (as the denominator of $AAV$ is the standard deviation which will be large) as mentioned by \cite{Silveira2010b,Silveira2010c}.

The results and analysis clearly suggest, as has been noted before by ~\cite{Silveira2010a}, that a hybrid approach consisting of {\footnotesize ASTUTE} and Kalman (or Wavelet) will capture most of the anomalies.
Importantly, Fig.~\ref{fig:Char} shows that the proposed M-SSA based approach is able to detect anomalies regardless of their location on
the feature properties map. M-SSA is able to detect significant temporal changes
in traffic as well as changes in the number of flows. M-SSA searches for correlation across flows properties ({\footnotesize ASTUTE} applies the same search concept between flows), while at the same time looking for temporal variation in a lag
window dimension of $\ell$. {\footnotesize ASTUTE} is limited to two
consecutive time bins.
\subsection{Detection Performance}
In order to evaluate the robustness and sensitivity of M-SSA we have
designed a simulation set up where we inject artificial anomalies
in real traces and measure the trade-off between the true positive
and false positives using ROC curves. One of the biggest challenges
in network anomaly detection systems, and which has limited their
widespread adoption, is the high false positive rate exhibited
by most existing techniques (see ~\cite{Ringberg2008,Axelsson2000}).

\subsubsection{Simulation}
\label{sec:simulation}

Our simulation is based on real trace data augmented with anomalous traffic injected in a similar fashion as in~\cite{Silveira2010a,Ringberg2008,Axelsson2000}. However and in addition to previous work, we build a simulation model which captures several distinctive characteristics of anomalies. We consider the distribution of time
between anomalies, duration,  magnitude (packet count for DoS attacks, number of flows for port scans, etc), and the anomaly type distribution (DoS, port scan, etc).

\begin{figure} [t!]
\centering
\includegraphics[width=0.8\textwidth]{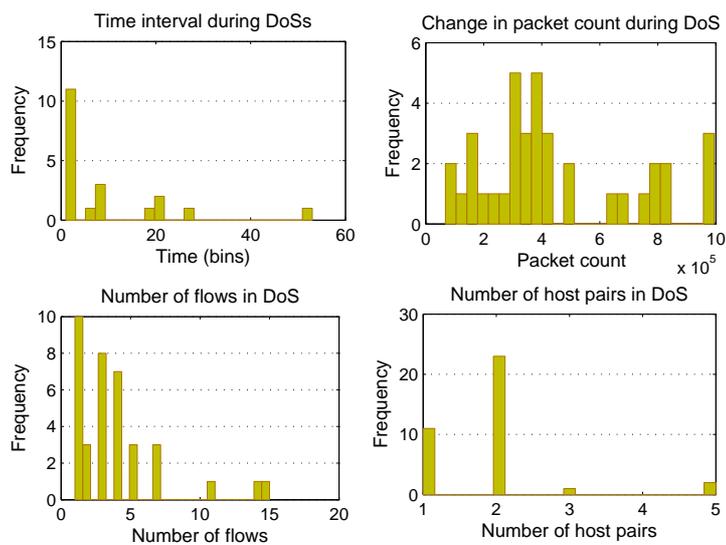}
\caption{Illustration of the distribution histograms used to simulate DoS attacks. Distribution histograms characterize the duration of attacks and size of attack (e.g. number of flows involved in the attack plus the change in the packet volume)}
\label{fig:PDF_dos}
\end{figure}

\begin{figure} [t!]
\centering
\includegraphics[width=0.8\textwidth]{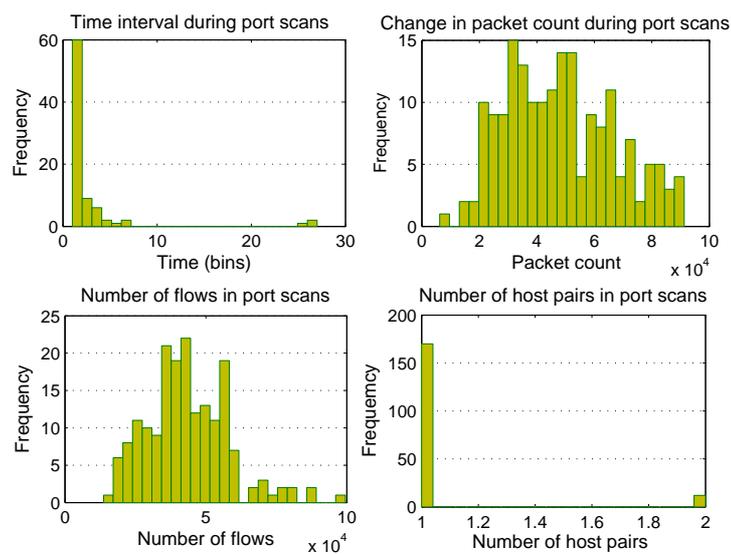}
\caption{Illustration of the distribution histograms used to simulate port scans. Distribution histograms characterize the duration of attacks and size of attack (e.g. number of flows involved in the attack plus the change in the packet volume)}
\label{fig:PDF_ps}
\vspace{-10pt}
\end{figure}

We first estimate the above parameters based on available observations in traffic traces. For example Fig.~\ref{fig:PDF_dos} and Fig.~\ref{fig:PDF_ps}
show the histograms of these property values for DoS attacks and port scans respectively, as observed in the Abilene trace. We start the simulation assuming a non-anomalous time bin and choose the next attack time, by sampling from the empirical probability distribution of the time between
anomalies. The anomaly type is then also chosen by sampling from the anomaly type distribution. At this point, a synthetic anomaly is generated by sampling from the anomaly duration and magnitude distribution, and injected into the synthetic trace. This process is repeated until the end of the simulation. The resulting trace therefore inherit the most significant statistical properties of the real data, e.g. the frequency of attacks and their magnitude.
\subsubsection{Results}
\begin{figure} [t!]
\centering
\includegraphics[width=0.85\textwidth]{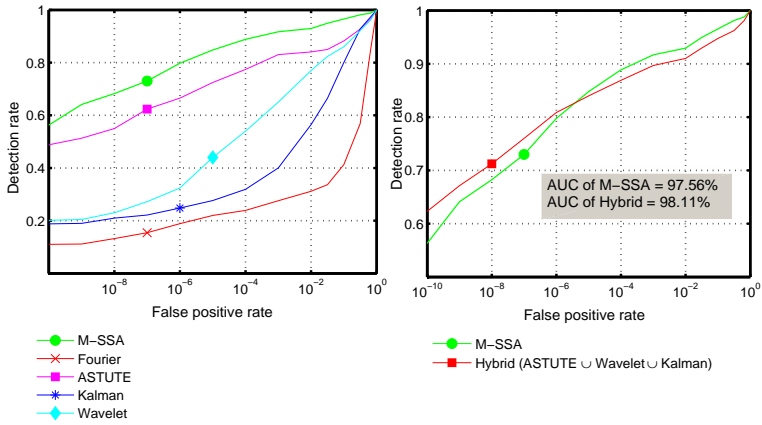}
\caption{ROC curves: M-SSA has a better detection rate than alternative techniques. A Hybrid system shows slightly better trade-off for a false positive rate less than $10^{-5}$.}
\label{fig:ROC2}
\end{figure}

The trade-off between false positive and true positive rate using the simulation data are captured using the ROC curve and are shown in Fig.~\ref{fig:ROC2}.
The simulation parameters for all algorithms are set as per Table~\ref{table:methods}. The ROC curves depicted in Fig.~\ref{fig:ROC2} show that M-SSA has higher true positive rate for a given false positive rate, compared with all other techniques. For example, for a false positive rate of 0.01\%, M-SSA detects 90\% of anomalies, whereas  Wavelet and {\footnotesize ASTUTE} only detect 77\% and 81\% respectively.
A Hybrid detector including Wavelets, Kalman and {\footnotesize ASTUTE} shows slightly better trade-off for a false positive rate less than $10^{-5}$ but M-SSA is better for the rest of interval. The Area Under Curve (AUC) which measures the overall performance of the detector has been shown in Fig.~\ref{fig:ROC2} (left).

\subsection{Configuration of Parameters}
We now evaluate the impact of the parameters: Lag Window Length ($\ell$),
the dimensionality $k$ of the projected space and the detection threshold $q_{\beta}$. 

\subsubsection{Lag window length ($\ell$)}

The key take away from the theoretical literature is that for an appropriate choice of $\ell$, the Hankel matrix will capture the appropriate dynamics of the underlying system (see \cite{Takens1981}. According to \cite{Takens1981,Broomhead1986a, Broomhead1986c} and \cite{Ghil2002}, the choice of $\ell$  must consider the trade-off between the maximum period (frequency) resolved and the statistical confidence of the result. A large value of $\ell$  will potentially better capture the long range trends but the size of the covariance matrix will be larger which will have to be estimated from a time series of effective length $n-\ell+ 1$.\\

\begin{figure}
        \centering
        \begin{subfigure}[b]{0.5\textwidth}
                \centering
                \includegraphics[width=\textwidth]{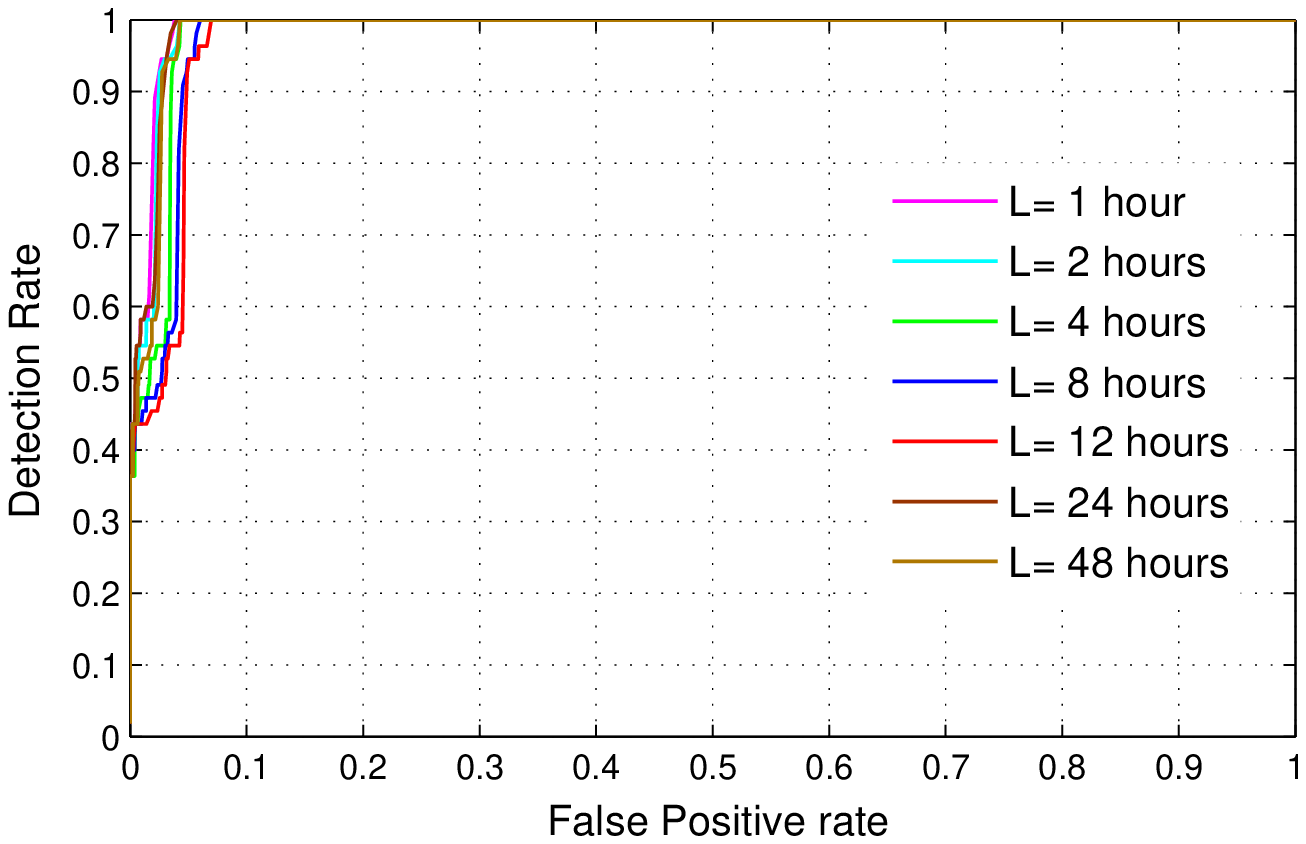}
                \caption{ $\begin{array}{c} \text{DoS detection performance} \\ \text{for different window length.}\end{array}$}
                \label{fig:DoS_RoC}
        \end{subfigure}%
        \begin{subfigure}[b]{0.5\textwidth}
                \centering
                \includegraphics[width=\textwidth]{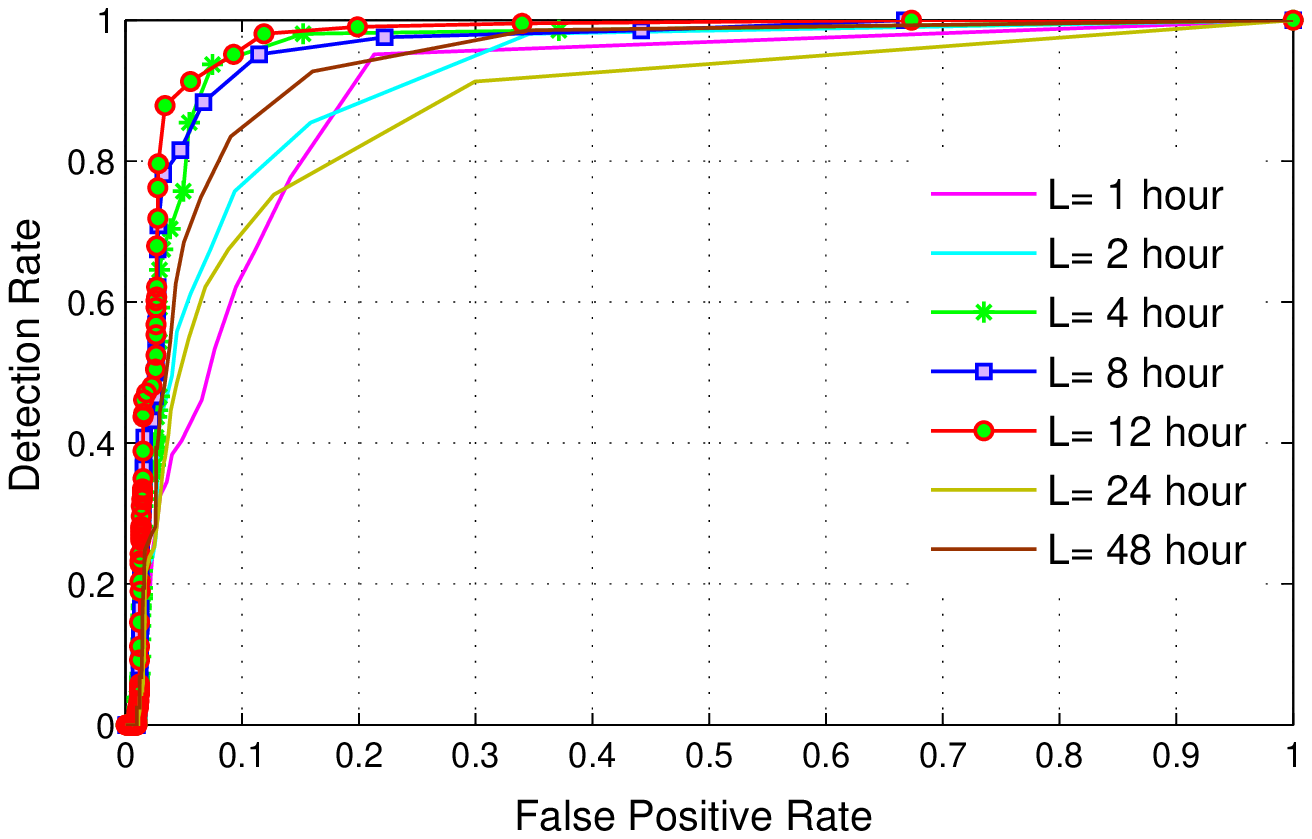}
                \caption{$\begin{array}{c} \text{Port scan  detection performance} \\ \text{for different window length.}\end{array}$}
                \label{fig:PtSn_RoC}
        \end{subfigure}

        \caption{The impact of window length $\ell$ on detecting DoS attacks and port scans. Notice that $\ell$ has almost no impact on on DoS detection but significant impact on port scan detection.}
        \label{fig:result}
\end{figure}

The choice of $\ell$ has a significant impact on detection performance of different anomalies.
DoS attacks and port scans  are emblematic of two types of deviations in network traffic. DoS attacks are characterized by large changes in a (relatively) small number of flows as the attacking hosts send a large number of small packets to deplete system resources in the attacked host (see Fig.~\ref{fig:PDF_dos} and Fig.~\ref{fig:Char}). Thus DoS like anomalies cause high temporal variation (within flows correlation) in the responsible flows and can be detected using techniques based on time series analysis.
Port scans, in the other hand, are characterized as small increases in a large number of flows (see Fig.~\ref{fig:PDF_ps} and Fig.~\ref{fig:Char}). This is required to detect for spatial correlation across flows (correlation between the flows) in order to find port scans.
We run an experiment to discuss the impact of window length on capturing temporal/spatial correlation, i.e. whithin/between flow correlation, of the traffic data. ROC curves in Fig.~\ref{fig:DoS_RoC} and Fig.~\ref{fig:PtSn_RoC} present DoS and port scan detection performance (separately) for varying window length. We describe the main findings learned from this experiment as follows. \\
\begin{itemize}
\item
It is clear that the detection of DoS is almost independent of the window length, see Fig.~\ref{fig:DoS_RoC}. This is expected as DoS attacks cause high correlation within flows (temporal variations) and this can be always captured even if the window length is zero, i.e. the common PCA is able to report them.  \\

\item
Across flows correlation is crucially dependent of window length as shown in Fig.~\ref{fig:PtSn_RoC}. Thus the choice of window size has significant impact on detecting port scans. When the window length is zero the correlation across the flows can not be captured. when the window length is large across flows correlation is suppressed. What is required is a localized window where deviation from normal correlation can be detected. According to the experiment, detecting port scans is improved for window length of $\ell=\{4,8,12\}$ (hours) while it is worsen for smaller/larger window length. 
\end{itemize}


\subsubsection{Grouping indices (\textit{k})}
\begin{figure} [t!]
\centering
\includegraphics[width=0.8\textwidth]{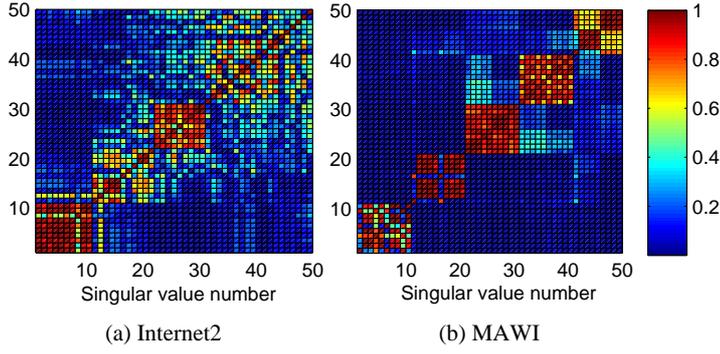}
\caption{Absolute values of w-correlation matrix plotted for the first 50 reconstructed components. The gaps in the scatter plot indicates how many components to select.}
\label{fig:Wcorr}
\vspace{-5pt}
\end{figure}

Another important parameter of M-SSA affecting results is the grouping indices, i.e. which components are grouped to provide the reconstructed data. The aim of our technique is to make a decomposition of the observed traffic into the sum
of underlying traffic system (can be a number of interpretable components such as a slowly varying trend, oscillatory components) and a structureless noise, as ${ Y = X + E }$. The decomposition of the series $Y$ into these two part is viable if the resulting additive
components $X$ and $E$ are approximately separable from each other. Suppose the the full reconstructed components are denoted by $V_{i}=Mz$ for $i=\{1,2,...,m \times \ell\}$.
To select which components to group, we compute the weighted correlation matrix (w-corr), where each element of the matrix $\rho _{ij}$ is defined as:
\[
\rho _{ij}=\frac{\textsl{cov}_w(V_i,V_j)}{\sigma_w(V_i)\sigma_w(V_j)}
\]
using:
\[
\sigma_w^2(V_i)=W'V_i^{'}V_i
\quad , \quad
\textsl{cov}_w(V_i,V_j)=W'V_i'V_j
\]
where  $w_t=min\{t,\ell,n-\ell\}$ for $t=\{1:n\}$ is the weighting vector.
If the absolute value of the w-correlations for two $V_i$ and $V_j$ is small (ideally zero), so the corresponding series are almost w-orthogonal and well separable. Fig.~\ref{fig:Wcorr} shows the absolute values of w-correlation for the
first 50 reconstructed components. This is a grade matrix plot from red (corresponding to 1) to blue (corresponding to 0), which shows both the separability and dominance of components with highest eigenvalues values. This plot is useful to select how many components to select in the reconstruction phase, as we only need to select the first $k$ components with the largest w-corr values.
From Fig.~\ref{fig:Wcorr}, we observe that the absolute value of the w-correlation for first 10 components are naturally grouped, a property that is observed for both the Abilene and MAWI datasets. We therefore suggest to use the first 10 components for the reconstruction when using M-SSA. So the $X=\sum_{i=1}^{i=10}V_i$ and residual space ${E = Y - X }$. In next section we will see that how the values of w-correlation can also be checked for adjusting the decision parameter ($\emph{q}_\beta$) so that a false positive rate can be met.

\subsubsection{Decision Variable ($\emph{q}_\beta$)}
For the decision threshold value (i.e., when to raise an alarm for any
anomaly investigating $E$ space), we use the variables proposed in previous studies (see~\cite{Lakhina2004a,Jackson1979,Jensen1972})in
network anomaly detection but we address the problem associated with this criteria as discussed by \cite{Ringberg2007}. The
threshold $\emph{q}_{\beta}$ is defined as
\[
\begin{array}{ll}
\emph{q}_\beta & =Q(\lambda_{k+1}:\lambda_{\ell \times m},\beta) \\ [2ex]
& =\phi_1[\frac{(1-\beta) \sqrt(2\phi_2h^2)}{\phi_1}+1+\frac{\phi_2h(h-1)}{\phi_1^2}]^{1/h}
\end{array}
\]
denotes the threshold for the 1$- \beta$ confidence level, corresponds to a false alarm rate of $\beta$, and
\[
h=1-\frac{2\phi_1\phi_3}{3\phi_2^2}, \quad  \phi_i=\sum_{j=k+1}^{\ell m} \lambda_i \quad \textit{for} \quad i=1,2,3.
  \]
Based on \cite{Jensen1972} the $Q$ in the above equation follows a gaussian distribution, and this convergence is robust even when the original data deviates from a gaussian distribution. \cite{Ringberg2007} had
questioned the robustness of the $Q$ metric - especially in the low
false positive regime.  \cite{Brauckhoff2009} have shown
that the main reason the metric is not robust is because the use of
standard PCA results in a residual which exhibits temporal
correlation. In principle the residual should correspond to noise and
be completely uncorrelated. Thus by ensuring that temporal correlation (in the case of KL transform) and spatio-temporal correlation (in the case of M-SSA) is captured by the model, the $Q$ metric is robust.

The w-correlation matrix computed above can help  verify if the residual space, given by ${E = Y - X }$ where $X$ is the reconstructed space, contains correlated elements or not. For example, the  w-correlation plot in Fig.~\ref{fig:Wcorr}
clearly shows that  that when $X$ is the space spanned by  $V_i$ for $i>10$,
the reconstructed elements are strongly w-orthogonal in both Abilene and MAWI traffic, resulting in uncorrelated residuals.

%% file: related.tex
\section{Related Work}
\label{sec:related}
Current network infrastructure is protected against malicious attacks
by signature-based Intrusion Detection Systems (IDS) (\cite{Roesch1999,Paxson1998}). However, it is well known that attackers can circumvent these systems by generating small modifications of known signatures.

In principle, anomaly-based detection systems (ADS) offer an attractive alterative to signature-based systems. ADS are based
on the notion of ''statistical normality'', and malicious events are
those that cause deviations from normal behavior. The major challenge
is to characterize normal traffic subject to the constraint
that network traffic exhibits non-stationary behavior.

Existing techniques for ADS are based on decomposition methods of network time series. For example \cite{Lakhina2004a,Lakhina2004b,Lakhina2005} has proposed the use of Principal
Component Analysis (PCA) for detection of network wide anomalies. \cite{Zhang2005}
has compared the use of Fourier, Wavelets and ARIMA methods for detection of
link anomalies and then have used $\ell_{1}$ optimization to recover the
origin-destination pairs which may have caused the link anomalies to appear.
Further refinements on PCA and state methods like Kalman Filtering have
been extensively investigated for first extracting the normal behavior
and then reporting deviations from normality as potential anomalies (see \cite{Barford2002,Lu2009,Zhang2005,brutlag2000,Krishnamurthy2003,Soule2005}).

The mathematical basis of Singular Spectrum Analysis (SSA) is
the celebrated result in nonlinear dynamics due to~\cite{Takens1981}.
Taken's theorem asserts that the latent non-linear dynamics governing
can be recovered using a delayed time embedding of the observable
time series. The first practical use of Taken's theorem for time series
analysis and the  connection with spectral methods like singular value decomposition (SVD) was first proposed by~\cite{Broomhead1986a,Broomhead1986b}.
Further application of the technique in climate and geophysical
time series analysis has been extensively investigated in~\cite{Vautard1992,Allen1996,Golyandina2010,Vautard1989,Ghil1991,Yiou1996,Ghil2002}.

%% file: conclusion.tex
\section{Conclusions}
\label{sec:conclusion}
In this paper we have proposed a unified and robust method for network
anomaly detection based on Multivariate Singular Spectrum Analysis (M-SSA).
As M-SSA can detect deviations from both spatial and temporal correlation
present in the data, it allows for the detection of both DoS and port scan attacks.
A DoS attack is an example of temporal deviation while a port scan attack
violates spatial correlation. Besides the use of M-SSA for network
anomaly detection, we have carried out a comprehensive evaluation and
compared M-SSA with other approaches based on wavelets, fourier analysis,
kalman filtering and the recently introduced ASTUTE method. We have also carried
out a rigorous analysis of the parameter configurations that accompany the use
of M-SSA and address some of the important issues that have been raised in the
networks community. Finally we have introduced a new labeled dataset from a large backbone
link between Japan and the United States.

%% file: hankelize.tex
The averaging over the diagonals $\mathrm{i+j=const}$ of the matrices $\mathbf{X}_{I_i}$ is called Hankelization. The purpose of diagonal averaging is to transform a matrix to the form of a Hankel matrix, which can be subsequently converted to a time series.
In other word, diagonal averaging maps matrices $\mathbf{X}_{I_i}$ into a time series. to be continued ...
Let Hankelization  operator $\mathcal{H}$ acting on any arbitrary matrix  to turn it into a Hankle matrix in an optimal way. By applying the Hankelization procedure to all matrix components of $\mathbf{X}_{I_i}$ the expansion will be:
\[
\mathbf{X}=\mathbb{X}_{I_1}+...+\mathbb{X}_{I_m}
\]
where $\mathbb{X}_{I_1} = \mathcal{H}X_{I_1}$.
Since all the matrices on the right-hand side of the expansion are Hankel matrices, each matrix uniquely specifies the time series and we thus obtain the decomposition of the original time series:
\[
\mathbb{X}_{I_1} \equiv Y_1(t) \quad ...  \quad \mathbb{X}_{I_m} \equiv Y_m(t)
\]

The complete original series is simply spume of the thus far obtained components.
 \[
 Y(t)=Y_1(t)+Y_2(t)+...+Y_m(t)
\]

%% file: template.bbl
\begin{thebibliography}{35}
\providecommand{\natexlab}[1]{#1}
\providecommand{\url}[1]{{#1}}
\providecommand{\urlprefix}{URL }
\expandafter\ifx\csname urlstyle\endcsname\relax
  \providecommand{\doi}[1]{DOI~\discretionary{}{}{}#1}\else
  \providecommand{\doi}{DOI~\discretionary{}{}{}\begingroup
  \urlstyle{rm}\Url}\fi
\providecommand{\eprint}[2][]{\url{#2}}

\bibitem[{Allen and Smith(1996)}]{Allen1996}
Allen MR, Smith LA (1996) {Monte Carlo SSA: Detecting irregular oscillations in
  the Presence of Colored Noise}. J Climate 9(12):3373--3404

\bibitem[{Axelsson(2000)}]{Axelsson2000}
Axelsson S (2000) The base-rate fallacy and the difficulty of intrusion
  detection. ACM Trans Inf Syst Secur 3(3):186--205

\bibitem[{Barford et~al(2002)Barford, Kline, Plonka, and Ron}]{Barford2002}
Barford P, Kline J, Plonka D, Ron A (2002) A signal analysis of network traffic
  anomalies. In: IMW'02 Proceedings of the 2nd ACM SIGCOMM Workshop on Internet
  measurment, pp 71--82

\bibitem[{Brauckhoff et~al(2009)Brauckhoff, Salamatian, and
  May}]{Brauckhoff2009}
Brauckhoff D, Salamatian K, May M (2009) {Applying PCA for traffic anomaly
  detection: Problems and solutions}. In: INFOCOM 2009, IEEE, IEEE, pp
  2866--2870

\bibitem[{Broomhead and King(1986{\natexlab{a}})}]{Broomhead1986a}
Broomhead D, King GP (1986{\natexlab{a}}) Extracting qualitative dynamics from
  experimental data. In: Physica D, Nonlinear Phenomena, vol~20, pp 217 -- 236

\bibitem[{Broomhead and King(1986{\natexlab{b}})}]{Broomhead1986c}
Broomhead D, King GP (1986{\natexlab{b}}) Extracting qualitative dynamics from
  experimental data. Physica D: Nonlinear Phenomena 20(2–3):217 -- 236

\bibitem[{Broomhead and King(1986{\natexlab{c}})}]{Broomhead1986b}
Broomhead DS, King GP (1986{\natexlab{c}}) {On the qualitative analysis of
  experimental dynamical systems}. Nonlinear Phenomena and Chaos S. Sarkar

\bibitem[{Brutlag(2000)}]{brutlag2000}
Brutlag JD (2000) Aberrant behavior detection in time series for network
  monitoring. In: Proceedings of the 14th USENIX conference on System
  administration, USENIX Association, pp 139--146

\bibitem[{Elsner and Tsonis(1996)}]{Elsner1996}
Elsner J, Tsonis A (1996) Singular Spectrum Analysis: A New Tool in Time Series
  Analysis. The language of science, Springer

\bibitem[{Ghil and Vautard(1991)}]{Ghil1991}
Ghil M, Vautard R (1991) {Interdecadal oscillations and the warming trend in
  global temperature time series}. Nature 350(6316):324--327

\bibitem[{Ghil et~al(2002)Ghil, Allen, Dettinger, Ide, Kondrashov, Mann,
  Robertson, Saunders, Tian, Varadi, and Yiou}]{Ghil2002}
Ghil M, Allen MR, Dettinger MD, Ide K, Kondrashov D, Mann ME, Robertson AW,
  Saunders A, Tian Y, Varadi F, Yiou P (2002) Advanced spectral methods for
  climatic time series. Reviews of Geophysics 40(1)

\bibitem[{Golyandina(2010)}]{Golyandina2011}
Golyandina N (2010) On the choice of parameters in singular spectrum analysis
  and related subspace-based methods. Statistics and Its Interface
  3(3):259--279

\bibitem[{Golyandina et~al(2010)Golyandina, Nekrutkin, and
  Zhigljavsky}]{Golyandina2010}
Golyandina N, Nekrutkin V, Zhigljavsky A (2010) Analysis of Time Series
  Structure: SSA and Related Techniques. Chapman \& Hall/CRC Monographs on
  Statistics \& Applied Probability, Taylor \& Francis

\bibitem[{Iokhvidov et~al(1982)Iokhvidov, Nicholson, Keaton, Beatty, Herrman,
  Griffith, Kosinski, and Pictures}]{Iokhvidov1982}
Iokhvidov IS, Nicholson J, Keaton D, Beatty W, Herrman E, Griffith T, Kosinski
  J, Pictures P (1982) Hankel and Toeplitz matrices and forms: algebraic
  theory. Birkh{\"a}user Boston

\bibitem[{Jackson and Mudholkar(1979)}]{Jackson1979}
Jackson JE, Mudholkar GS (1979) {Control Procedures for Residuals Associated
  with Principal Component Analysis}. Technometrics 21(3):341--349

\bibitem[{Jensen and Solomon(1972)}]{Jensen1972}
Jensen DR, Solomon H (1972) A gaussian approximation to the distribution of a
  definite quadratic form. Journal of the American Statistical Association
  67(340):898--902

\bibitem[{Juang and Pappa(1985)}]{Juang1985}
Juang J, Pappa R (1985) Eigensystem realization algorithm for modal parameter
  identification and model reduction. Journal of Guidance 8(5):620

\bibitem[{Krishnamurthy et~al(2003)Krishnamurthy, Sen, Zhang, and
  Chen}]{Krishnamurthy2003}
Krishnamurthy B, Sen S, Zhang Y, Chen Y (2003) Sketch-based change detection:
  methods, evaluation, and applications. In: Proceedings of the 3rd ACM SIGCOMM
  conference on Internet measurement, ACM, New York, NY, USA, IMC '03, pp
  234--247

\bibitem[{Lakhina et~al(2004{\natexlab{a}})Lakhina, Crovella, and
  Diot}]{Lakhina2004b}
Lakhina A, Crovella M, Diot C (2004{\natexlab{a}}) Characterization of
  network-wide anomalies in traffic flows. In: Proceedings of the 4th ACM
  SIGCOMM conference on Internet measurement, vol~35, pp 201--206

\bibitem[{Lakhina et~al(2004{\natexlab{b}})Lakhina, Crovella, and
  Diot}]{Lakhina2004a}
Lakhina A, Crovella M, Diot C (2004{\natexlab{b}}) Diagnosing network-wide
  traffic anomalies. SIGCOMM Comput Commun Rev 34(4):219--230

\bibitem[{Lakhina et~al(2005)Lakhina, Crovella, and Diot}]{Lakhina2005}
Lakhina A, Crovella M, Diot C (2005) Mining anomalies using traffic feature
  distributions. In: SIGCOMM Comput. Commun. Rev., vol~35, pp 217--228

\bibitem[{Lu and Ghorbani(2009)}]{Lu2009}
Lu W, Ghorbani AA (2009) Network anomaly detection based on wavelet analysis.
  EURASIP Journal on Advances in Signal Processing - Special issue on signal
  processing applications in network intrusion detection systems 2009:4:1--4:16

\bibitem[{Paxson(1998)}]{Paxson1998}
Paxson V (1998) Bro: a system for detecting network intruders in real-time. In:
  Proceedings of the 7th conference on USENIX Security Symposium - Volume 7,
  USENIX Association, SSYM'98, pp 3--3

\bibitem[{Ringberg et~al(2007)Ringberg, Soule, Rexford, and
  Diot}]{Ringberg2007}
Ringberg H, Soule A, Rexford J, Diot C (2007) Sensitivity of pca for traffic
  anomaly detection. SIGMETRICS Perform Eval Rev 35(1):109--120

\bibitem[{Ringberg et~al(2008)Ringberg, Roughan, and Rexford}]{Ringberg2008}
Ringberg H, Roughan M, Rexford J (2008) The need for simulation in evaluating
  anomaly detectors. SIGCOMM Comput Commun Rev 38(1):55--59

\bibitem[{Roesch(1999)}]{Roesch1999}
Roesch M (1999) Snort - lightweight intrusion detection for networks. In:
  Proceedings of the 13th USENIX conference on System administration, USENIX
  Association, LISA '99, pp 229--238

\bibitem[{Silveira and Diot(2010)}]{Silveira2010c}
Silveira F, Diot C (2010) Urca: pulling out anomalies by their root causes. In:
  Proceedings of the 29th IEEE INFOCOM 2010 conference, pp 722--730

\bibitem[{Silveira et~al(2010{\natexlab{a}})Silveira, Diot, Taft, and
  Govindan}]{Silveira2010a}
Silveira F, Diot C, Taft N, Govindan R (2010{\natexlab{a}}) Astute: detecting a
  different class of traffic anomalies. In: Proceedings of the ACM SIGCOMM 2010
  conference, pp 267--278

\bibitem[{Silveira et~al(2010{\natexlab{b}})Silveira, Diot, Taft, and
  Govindan}]{Silveira2010b}
Silveira F, Diot C, Taft N, Govindan R (2010{\natexlab{b}}) Detecting traffic
  anomalies using an equilibrium property. In: Proceedings of the ACM
  SIGMETRICS 2010 conference, pp 377--378

\bibitem[{Soule et~al(2005)Soule, Salamatian, and Taft}]{Soule2005}
Soule A, Salamatian K, Taft N (2005) Combining filtering and statistical
  methods for anomaly detection. In: IMC '05 Proceedings of the 5th ACM SIGCOMM
  conference on Internet Measurement, IMC '05, pp 31--31

\bibitem[{Takens et~al(1981)Takens, Rand, and Young}]{Takens1981}
Takens F, Rand D, Young LS (1981) Detecting strange attractors in turbulence.
  In: Lecture Notes in Mathematics, vol 898, pp 366 -- 381

\bibitem[{Vautard and Ghil(1989)}]{Vautard1989}
Vautard R, Ghil M (1989) Singular spectrum analysis in nonlinear dynamics, with
  applications to paleoclimatic time series. In: Physica D, Nonlinear
  Phenomena, vol~35, pp 395 -- 424

\bibitem[{Vautard et~al(1992)Vautard, Yiou, and Ghil}]{Vautard1992}
Vautard R, Yiou P, Ghil M (1992) Singular-spectrum analysis: a toolkit for
  short, noisy chaotic signals. Phys D 58(1-4):95--126

\bibitem[{Yiou et~al(1996)Yiou, Baert, and Loutre}]{Yiou1996}
Yiou P, Baert E, Loutre M (1996) Spectral analysis of climate data. Surveys in
  Geophysics 17(6):619--663

\bibitem[{Zhang et~al(2005)Zhang, Ge, Greenberg, and Roughan}]{Zhang2005}
Zhang Y, Ge Z, Greenberg A, Roughan M (2005) Network anomography. In:
  Proceedings of the 5th ACM SIGCOMM conference on Internet Measurement, IMC
  '05

\end{thebibliography}
